\documentclass[letterpaper, 10 pt, conference]{ieeeconf}  

\IEEEoverridecommandlockouts                              

\usepackage{graphics} 
\usepackage{epsfig} 
\usepackage{mathptmx} 
\usepackage{times} 
\usepackage{amsmath} 
\usepackage{amssymb}  
\usepackage{amsthm}
\usepackage[noend]{algpseudocode}
\usepackage{algorithm}
\usepackage{xcolor}
\usepackage{cite}
\usepackage{bm}
\usepackage{balance}

\theoremstyle{definition}

\theoremstyle{definition}
\newtheorem{problem}{Problem}
\theoremstyle{definition}

\DeclareMathOperator*{\argmax}{arg\; max}     

\title{\LARGE \bf
Path-Based Sensors: Will the Knowledge of Correlation in Random Variables Accelerate Information Gathering?
}

\author{Alkesh K. Srivastava$^{1}$, George P. Kontoudis$^{1}$, Donald Sofge$^{2}$, and Michael Otte$^{1}$
\thanks{$^{1}$Alkesh K. Srivastava, George P. Kontoudis,  Michael Otte are with the Department of Aerospace Engineering, University of Maryland, College Park, MD, USA. Emails: 
        \{alkesh, kont, otte\}@umd.edu}%
\thanks{$^{2}$Donald Sofge is with the U.S. Naval Research Laboratory,
        Washington, DC, USA. E-mail: 
        donald.sofge@nrl.navy.mil}%
\thanks{This work is supported by the Maryland Robotics Center and the Office of Naval Research (ONR) via grant N$0001420$WX$01827$~and~N$00014$-$20$-$1$-$2712$. The views, positions, and conclusions contained in this document are solely those of the authors and do not explicitly represent those of ONR.}
}

\begin{document}

\maketitle
\thispagestyle{empty}
\pagestyle{empty}

\begin{abstract}
Effective communication is crucial for deploying robots in mission-specific tasks, but inadequate or unreliable communication can greatly reduce mission efficacy, for example in search and rescue missions where communication-denied conditions may occur. 
In such missions, robots are deployed to locate targets, such as human survivors, but they might get trapped at hazardous locations, such as in a trapping pit or by debris. Thus, the information the robot collected is lost owing to the lack of communication. In our prior work, we developed the notion of a path-based sensor. A path-based sensor detects whether or not an event has occurred along a particular path, but it does not provide the exact location of the event. Such path-based sensor observations are well-suited to communication-denied environments, and various studies have explored methods to improve information gathering in such settings. In some missions it is typical for target elements to be in close proximity to hazardous factors that hinder the information-gathering process. In this study, we examine a similar scenario and conduct experiments to determine if additional knowledge about the correlation between hazards and targets improves the efficiency of information gathering. To incorporate this knowledge, we utilize a Bayesian network representation of domain knowledge and develop an algorithm based on this representation. Our empirical investigation reveals that such additional information on correlation is beneficial only in environments with moderate hazard lethality, suggesting that while knowledge of correlation helps, further research and development is necessary for optimal outcomes.

\end{abstract}

\section{Introduction}
\label{sec:introduction}
\begin{figure}[!t]
    \centering
    \includegraphics[width=\columnwidth]{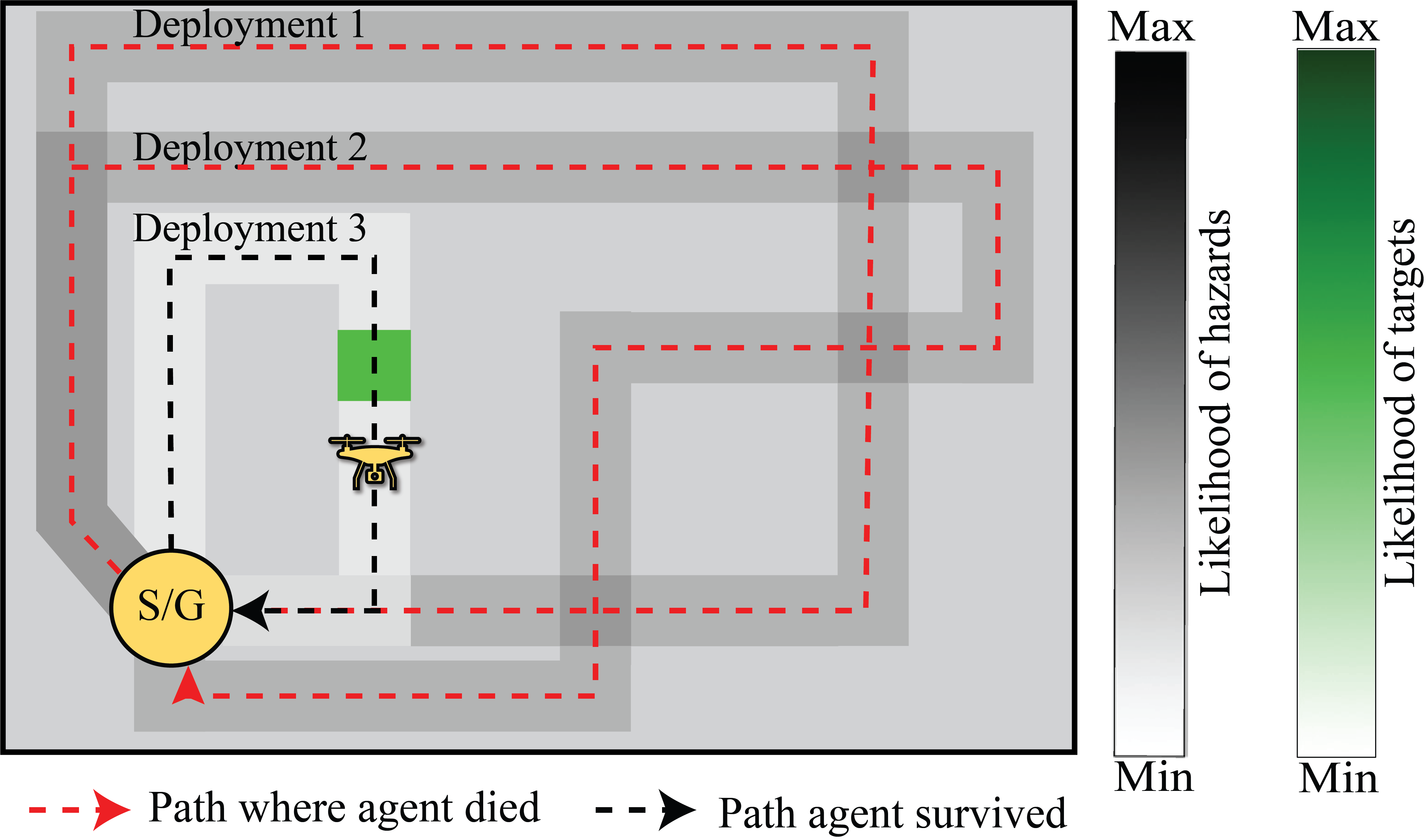}
    \caption{Estimation of hazardous locations and target elements using path-based sensor observations by sequential deployment of agents when inter-agent communication is prohibited. The red line depicts a path an agent attempted to traverse, but was destroyed somewhere along the path and never reached the goal position, while the black line shows the path along which the agent survived. In deployments $1$ and $2$, the agents never reached the goal; as a result, there is a greater likelihood of hazard in the overlapping zones. In contrast, an agent survived the third deployment; consequently, the overlapping zone has a lower probability of hazard. The agent is capable of detecting the precise location of target elements with a certain degree of confidence. Consequently, the greened area is more likely to contain a target with a higher probability.}
    \label{fig:concept_image}
    
\end{figure}

In recent years there has been increasing interest in the development of efficient and robust algorithms for information gathering in hazardous and communication-denied environments. These environments pose significant challenges to agents as they face risks from hazards while also having limited or no communication capabilities with a centralized server. In this context, path-based sensor observations have emerged as a common and useful means for gathering information about the environment. However, these sensors detect whether an event has occurred along a particular path, without providing the exact location of the event.

Recent work by \textit{Srivastava et al.}~\cite{Srivastava.etal.RAL23} has shown that Bayesian network-based approaches have promising results in information-theoretic planning and belief updates using path-based sensor observations. Building on this work, this paper focuses on improving the efficiency and robustness of information gathering in communication-denied environments with 
hazardous target elements 
by leveraging the additional knowledge of correlation between them, referred to as $\kappa$-correlation.

We present a new algorithm called $\kappa$-BNITP that uses a specialized Bayesian network to leverage the domain knowledge of hazard-target connectivity. Our numerical experiments show that $\kappa$-BNITP outperforms existing approaches, namely \textsc{relaxed-ITP}~\cite{Otte.Sofge.TASE21} and \textsc{relaxed-BNITP}~\cite{Srivastava.etal.RAL23}, in terms of information entropy reduction at moderate hazard lethality levels. However, we note that the algorithm's performance is reduced at higher and lower hazard lethality levels due to: i) the unreliable information obtained through path-based sensors; and ii) the Bayesian network approach's reliance on the accuracy of domain knowledge.

Overall, the incorporation of additional domain knowledge such as $\kappa$-correlation can improve the efficacy of information gathering in communication-denied environments. Future research is needed to develop more accurate models and approaches to estimate the environment.
\section{Related Work}
\label{sec:rw}
Efficient algorithms for information gathering in hazardous and communication-denied environments have gained significant interest in recent years. Path-based sensor observations are commonly used in such scenarios, and various studies have been conducted to explore information gathering using path-based sensors~\cite{Otte.etal.SPAR20B, Otte.Sofge.TASE21, Srivastava.etal.DARS22, Srivastava.etal.RAL23}. Bayesian frameworks have shown promise in information-theoretic planning and target estimation in robotics~\cite{tomo}.

Information theory~\cite{shannon1948mathematical} provides the foundation for information-theoretic planners, and a connection between Bayesian inference and information theory has been established~\cite{oladyshkin2019connection}. Methods that combine Bayesian frameworks and information theory to solve inverse modeling problems have been surveyed~\cite{mohammad2015entropy}. Mutual information, a measure of relative information derived from information theory, has been widely used in robotics~\cite{charrow2014approximate, grocholsky2002information, bourgault2002information}. \emph{Julian et al.}~\cite{julian2012distributed} derived the gradient of mutual information and demonstrated that information entropy would approach zero under a multi-agent gradient ascent control strategy. However, their work assumes explicit observations of hazards, while we consider implicit observations from path-based sensors.

Probabilistic graphical models have been proposed as a way to represent conditional dependencies among involved random variables~\cite{pearl1988probabilistic, koller2009probabilistic}. A probabilistic graphical model-based method is used for target tracking in \cite{uney2007graphical, uney2008target}, while in \cite{Srivastava.etal.RAL23}, a Bayesian network approach is used to estimate the origin of path-based sensor realizations. However, the study conducted in \cite{Srivastava.etal.RAL23} did not consider the inherent correlation between hazards and targets in the environment. In this paper, we propose a novel method that incorporates this correlation, which we define as $\kappa$-correlation. We also present an empirical study of the efficacy of our proposed method.
\section{Problem Formulation}
\label{sec:problem_formulation}
Consider a search space $\mathbf{S}\in \mathbb{R}^2$ consisting of $a \times b$ discrete cells, wherein each cell may be empty, contain hazardous elements (denoted by ${\mathbf{Z}}$) that can potentially lead to the destruction of the agent, contain elements of interest known as targets (denoted by $\mathbf{X}$), or contain both hazards and targets. The presence of hazards in a cell is denoted by $Z = 1$ and absence by $Z = 0$, while the presence of a target is denoted by $X = 1$ and absence by $X = 0$. For each discrete cell, we assume that there exists a correlation $\kappa$ between hazards $Z$ and targets $X$, which is defined as the likelihood of the existence of a target given there is a hazard i.e., $\kappa = \mathbb{P}(X = 1 | Z = 1)$.

A team of $M$ agents is deployed sequentially in the search space. An agent can visit up to $l$~cells during a deployment such that the deployment starts and ends at a fixed cell, called the base station~$d$, forming the path $\zeta_{m} = \langle d, c_{1},c_{2},\ldots,c_{l-2}, d\rangle.$ If the agent $m$ is destroyed {anywhere} along the path $\zeta_m$  then the path-based sensor is triggered ($\Theta = 1$), and if the agent $m$ survives the path $\zeta_m$ then we consider that the path-based sensor is not triggered ($\Theta = 0$). The path-based sensor is subject to false-positive and false-negative triggering, where false-positive corresponds to faulty or malfunctioning robots that get destroyed regardless of the presence of a hazard, and false-negative corresponds to robots surviving a cell despite having a hazard. The presence of a target is recorded by a noisy sensor that may also report a false positive or a false negative observation of the target. Unlike the path-based sensor, the target sensor reports the exact location of the target observation ($Y = 1$) along the path, if a robot survives it.

\begin{problem}
Given a team of $M$ agents sequentially deployed in a communication-denied environment, the task of an agent $m$ is to find a path $\zeta^*_{m}$ that maximizes the expected information gained about targets $X$ and hazards $Z$,
\label{ps:problem}
\end{problem}
\begin{equation}
\zeta^*_{m} = \argmax_{\zeta}{\{I(X_t;Y_{t+1},\ldots,Y_{t+l}|\Theta_{\zeta}, \kappa) + I(Z_t;\Theta_{\zeta | \kappa})\\}\}.
\end{equation}

\section{Incorporating $\kappa$-correlation with Bayesian Networks in Path-Based Sensors}
\label{sec:approach}
In this section, we introduce a \emph{Bayesian network} approach to incorporate the additional domain knowledge of $\kappa$-correlation to study its effect on information gathering. In our prior works, we extensively study the problem of maximizing information gathering using path-based sensors~\cite{Otte.etal.SPAR20B, Otte.Sofge.TASE21, Srivastava.etal.DARS22, Srivastava.etal.RAL23}. In \cite{Otte.etal.SPAR20B, Otte.Sofge.TASE21, Srivastava.etal.DARS22}, an analogy of multi-universe is used to infer from path-based sensor observations. If an agent traverses a path with \textit{l}~discrete cells then \textit{l} separate universes are considered where in each universe the path-based sensor trips at a particular cell from the set of \textit{l}~mutually exclusive cells, and then the final belief map is computed by integrating the resulting \textit{l} belief maps weighted by their relative probability of occurring. In contrast, \cite{Srivastava.etal.RAL23} propose a methodology that uses a Bayesian network to incorporate the multi-universe idea of~\cite{Otte.etal.SPAR20B, Otte.Sofge.TASE21, Srivastava.etal.DARS22} into a single joint distribution.

\begin{figure}[!t]
    \centering
    \includegraphics[width=0.6\columnwidth]{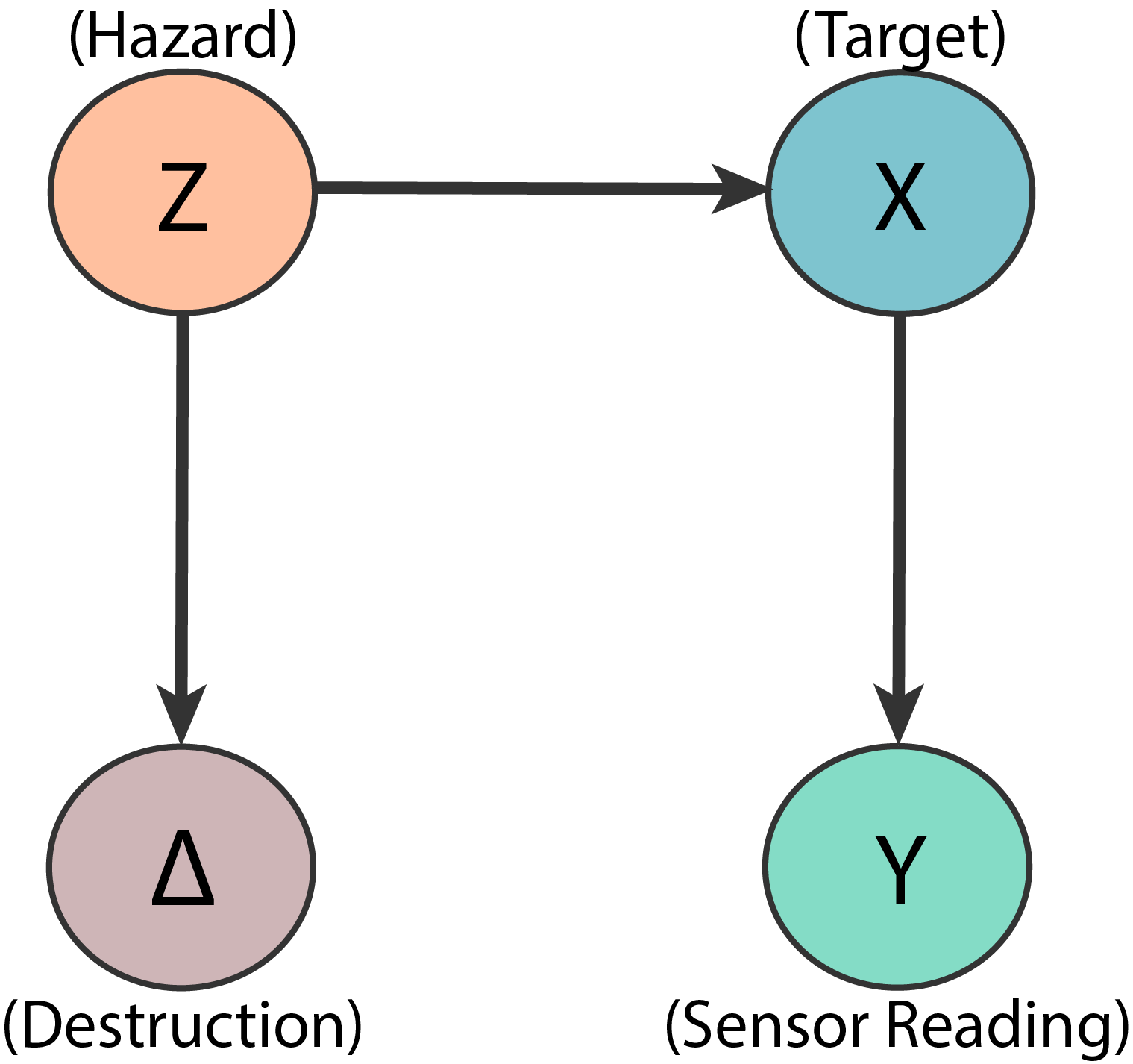}
    \caption{Bayesian Network representation of domain knowledge showing conditional dependencies in a given cell. Hazard $Z$ is the reason behind an agent's destruction $\Delta$, while the sensor reading $Y$ is dependent on the presence of target $X$. The Bayesian Network also incorporates $\kappa$-correlation, meaning that the presence of a target is dependent on the presence of hazards by some likelihood.}
    \label{fig:bayesian_network_1} 
\end{figure}
A Bayesian network is a probabilistic graphical model that represents a set of random variables and their conditional dependencies using a directed acyclic graph. In a Bayesian network, each node in the graph represents a random variable, and each edge represents a conditional dependency between the connected nodes. For a given discrete cell, the domain knowledge can be represented as the Bayesian network shown in Fig.~\ref{fig:bayesian_network_1}. In this study, the use of a Bayesian network provides the flexibility to incorporate $\kappa$-correlation by connecting an edge between the hazard node and target node, as in  Fig.~\ref{fig:bayesian_network_1}. The joint distribution of this Bayesian network is,
\begin{equation}
    \mathbb{P}(Z=z,X=x,\Delta=\delta,Y=y) =\mathbb{P}(z)\mathbb{P}(x|z)\mathbb{P}(\delta|z)\mathbb{P}(y|x). 
\end{equation}
\begin{figure}[!t]
    \centering
    \includegraphics[width=.85\columnwidth]{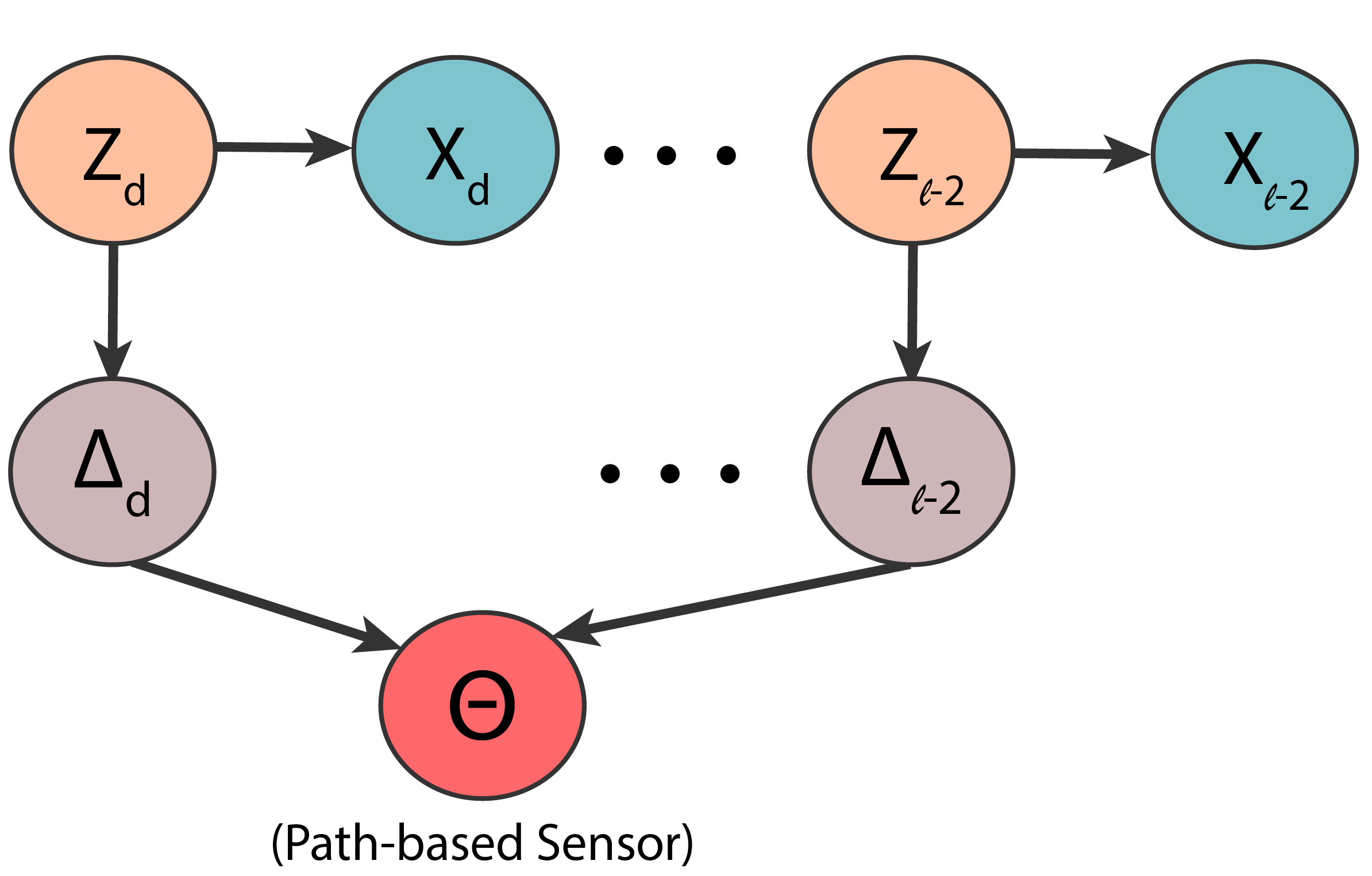}
    \caption{Bayesian Network representation of domain knowledge showing conditional dependencies in a given path~$\zeta_m$. Hazard $Z$ is the reason behind an agent's destruction $\Delta$ with incorporated $\kappa$-correlation. 
    The path-based sensor triggering $\Theta$ along the path~$\zeta_m$ is dependent on the destruction random variables $\Delta_d,\ldots,\Delta_{l-2}$}.
    \label{fig:BN_Layers} 
\end{figure}
Consider the environment described in Section~\ref{sec:problem_formulation}. If an agent traverses a path $\zeta_{d}=\langle d, C_1, \ldots C_{l-2}, d\rangle$ and never reaches its designated goal cell $C_l$. The problem can be modeled in a Bayesian network, as shown in Fig.~\ref{fig:BN_Layers}. Note that, in Fig.~\ref{fig:BN_Layers} the random variable of target sensor reading $\mathbf{Y}$ is excluded in the Bayesian network as it is irrelevant in the event of path-based sensor triggering. The resulting joint distribution is equal to the sum of the product for all possible distributions, 
\begin{equation}
\label{eq:BN_joint_distribution}
\mathbb{P}(Z, X, \Delta, \Theta)  = \sum_{z,x,\delta,\theta \in \{0, 1\}}\ \prod_{i = 1}^l\mathbb{P}(Z_i = z, X_i = x, \Delta_i = \delta, \Theta = \theta),
\end{equation}
where the joint distribution for the $i^{th}$ cell is,
\begin{align}
\nonumber
\mathbb{P}(Z_i, X_i, \Delta_i, \Theta) =\\\mathbb{P}(\Theta = \theta | \boldsymbol{\Delta}_{\zeta} =\boldsymbol{\delta})\mathbb{P}(\Delta_i = \delta | Z_i = z)\mathbb{P}(X_i = x | Z_i = z)\mathbb{P}(Z_i = z),
\end{align}
where $\mathbb{P}(\Theta = \theta | \boldsymbol{\Delta}_{\zeta}=\boldsymbol{\delta})$ is the likelihood of a path-based sensor trigger $\Theta = \theta$  along the path $\zeta$, $\mathbb{P}(\Delta_i = \delta | Z_i = z)$ is the likelihood of destruction $\Delta_i$ given the existence of hazard $Z = z$, $\mathbb{P}(X_i = x | Z_i = z)$ is the $\kappa$-correlation or the likelihood of the existence of target $X$ given hazard $Z$, and $\mathbb{P}(Z_i = z)$, is the prior belief of the hazard $Z_i$ at cell $C_i$ of path $\zeta$. The term $\mathbb{P}(\Theta = \theta | \boldsymbol{\Delta}_{\zeta}=\boldsymbol{\delta})$ is defined manually to incorporate the sequential dependency of a path $\zeta_m$ and estimates the likelihood of tripping the path-based sensor given a specific permutation of $l$ plausible causes as

\begin{equation}\label{eq:inference}
\begin{aligned}
\mathbb{P}(\Theta = 1|\boldsymbol{\Delta} = \boldsymbol{\delta}) = \begin{cases}
\prod_{i = 1}^{j} \mathbb{P}(\Delta_i = \delta_i), &\textrm{if} \ \sum_{i = 1}^l \delta_i \leq 1\\
0, &\textrm{if}\  \sum_{i = 1}^l \delta_i > 1,
\end{cases}
\end{aligned}
\end{equation} 
where $j$ is the index of the cell $C_j$ at which $\Delta_j = 1$. Therefore, this likelihood function takes into account the survival of the agent through the preceding $j - 1$ cells.

Note that, the Bayesian network representation in Fig.~\ref{fig:BN_Layers} and the likelihood function definition in~(\ref{eq:inference}) assumes that the path $\zeta_m$ does not include repeated cells. In other words, $\zeta_m~=~\langle d, c_{1},c_{2},\ldots,c_{l-2}, d\rangle$ is the same as the ordered set $\{ d, c_{1},c_{2},\ldots,c_{l-2}, d\}$. However, we allow an agent to traverse a path with repeated cells by altering the likelihood function~(\ref{eq:inference}) to include each possible instance of a repeated cell based on the path~$\zeta_m$. 

The calculation of the posterior belief map in the given environment is dependent on whether or not the path-based sensor was triggered, as this condition influences the Bayesian network and, consequently, the mathematical framework used for calculation.

\subsection{Case 1: Path-based sensor is not triggered}
When the path-based sensor is not triggered, the Bayesian network framework does not incorporate the random variable~$\Theta$ as it is deemed irrelevant, and the computation of the posterior belief map is performed independently for each discrete cell along the agent's path. In this scenario, the Bayesian network shown in Fig.~\ref{fig:bayesian_network_1} is utilized for computation. 

For a cell $i$ in path $\zeta_m$, the posterior belief of hazard $Z_i$ in that cell given sensor observation $Y_i = y$ is given by
\begin{multline}
    \mathbb{P}(Z_i = 1|Y_i = y, \Delta_i = 0) \propto\\ \sum_{x\in\{0,1\}}[\mathbb{P}(X_i = x|Z_i = 1)\mathbb{P}(\Delta_i = 0 | Z_i = 1)\\\times\mathbb{P}(Y_i = y | X_i =x)\mathbb{P}(Z_i = 1)].
    \label{eq:no_z}
\end{multline}
Similarly, the posterior belief of target $X_i$ is
\begin{multline}
    \mathbb{P}(X_i = 1|Y_i = y, \Delta_i = 0) \propto\\ \sum_{z\in\{0,1\}}[\mathbb{P}(X_i = 1|Z_i = z)\mathbb{P}(\Delta_i = 0 | Z_i = z)\\\times\mathbb{P}(Y_i = y | X_i =1)\mathbb{P}(Z_i = z)].
    \label{eq:no_x}
\end{multline}
The $\kappa$-correlation is also calculated and updated for each cell based on observations. The update rule follows
\begin{multline}
    \mathbb{P}(X_i = 1|Z_i = 1, Y_i = y, \Delta_i = 0) \propto\\ \mathbb{P}(X_i = 1|Z_i = 1)\mathbb{P}(\Delta_i = 0 | Z_i = 1)\mathbb{P}(Y_i = y | X_i =1)\mathbb{P}(Z_i = 1).
    \label{eq:no_kappa}
\end{multline}
The posterior probabilities are normalized in the standard Bayesian manner.
\subsection{Case 2: Path-based sensor is triggered}
When the path-based sensor is triggered, it is unknown where the agent may have been destroyed. Therefore, the random variable $\Theta$ is relevant and the Bayesian network shown in Fig.~\ref{fig:BN_Layers} is used for calculations. The calculation of the posterior belief map is similar to~\cite{Srivastava.etal.RAL23} but with the additional consideration of $\kappa$-correlation. The update rules based on the joint distribution shown in equation~(\ref{eq:BN_joint_distribution}). The posterior belief of hazard $Z_i$ in cell $i$ is given by 
\begin{multline}
\mathbb{P}(Z_i = 1 | \Theta = 1) \propto\\
\sum_{x\in\{0,1\},\boldsymbol{\delta}\in\boldsymbol{\Omega}}[\mathbb{P}(\Theta = 1 | \boldsymbol{\Delta}_{\zeta} =\boldsymbol{\delta})\mathbb{P}(\Delta_i = \delta | Z_i = 1)\\\times\mathbb{P}(X_i = x | Z_i = 1)\mathbb{P}(Z_i = 1)],
\label{eq:pbs_z}
\end{multline}
the posterior belief of target $X_i$ in cell $i$ is given by
\begin{multline}
\mathbb{P}(X_i = 1 | \Theta = 1) \propto\\
\sum_{z\in\{0,1\},\boldsymbol{\delta}\in\boldsymbol{\Omega}}[\mathbb{P}(\Theta = 1 | \boldsymbol{\Delta}_{\zeta} =\boldsymbol{\delta})\mathbb{P}(\Delta_i = \delta | Z_i = z)\\\times\mathbb{P}(X_i = 1 | Z_i = z)\mathbb{P}(Z_i = z)],
\label{eq:pbs_x}
\end{multline}
and the $\kappa$-correlation for cell $i$ is calculated as
\begin{multline}
\mathbb{P}(X_i = 1 | Z_i = 1, \Theta = 1) \propto\\
\sum_{x\in\{0,1\},\boldsymbol{\delta}\in\boldsymbol{\Omega}}[\mathbb{P}(\Theta = 1 | \boldsymbol{\Delta}_{\zeta} =\boldsymbol{\delta})\mathbb{P}(\Delta_i = \delta | Z_i = 1)\\\times\mathbb{P}(X_i = 1 | Z_i = 1)\mathbb{P}(Z_i = 1)].
\label{eq:pbs_kappa}
\end{multline}
In equation~(\ref{eq:pbs_z}),~(\ref{eq:pbs_x}),~and~(\ref{eq:pbs_kappa}) $\boldsymbol{\Omega}$ refers to the space of possible permutation of $\boldsymbol{\Delta_\zeta}$.

\section{Algorithms}
\label{sec:algorithms}

In this section, to address Problem~\ref{ps:problem} we propose a $\kappa$-correlation incorporating Bayesian network-based information-theoretic planner~(BNITP), called \textsc{$\kappa$-BNITP}. The proposed algorithm is similar to~\cite[Algorithm 1]{Srivastava.etal.RAL23} but incorporates $\kappa$-correlation as described in Section~\ref{sec:approach}.
\begin{algorithm}[!t]
\caption{\textsc{$\kappa$-BNITP}}\label{alg:mutual_info_planning}
\textbf{Inputs}: prior hazard belief map $Z^{(0)}$, prior target belief map $X^{(0)}$, prior $\kappa$-correlation belief map $\kappa^{(0)}$, search space $\mathrm{S}$, number of agents $M$\\
\textbf{Output}: posterior hazard belief map $Z^{(m)}$, posterior target belief map $X^{(m)}$, posterior $\kappa$-correlation belief map $\kappa^{(m)}$
\begin{algorithmic}[1]
\For{$m = 1,\ldots,M$}
\State $\mathbf{B}^{(m-1)}$ = $(Z^{(m-1)},X^{(m-1)},\kappa^{(m-1)})$\newline\Comment{Store Prior Belief Maps as a tuple in $\mathbf{B}^{(m-1)}$}
\State $\zeta_{m} \gets {\color{blue}\texttt{calculateBNPath}}(Z^{(m-1)},X^{(m-1)},\kappa^{(m-1)}, S)$
\State observe $\Theta$ and $Y^{(m)}$ after agent traverses path $\zeta_{m}$
\If{$\Theta = 0$}
\State $Z^{(m)} = {\color{blue}\texttt{Z-noPBS}}(Y^{(m)}, \mathbf{B}^{(m-1)}, \zeta_{m}, \Theta)$~~~~~~~~~~~~~~{\eqref{eq:no_z}}
\State $X^{(m)} = {\color{blue}\texttt{X-noPBS}}(Y^{(m)}, \mathbf{B}^{(m-1)}, \zeta_{m}, \Theta)$~~~~~~~~~~~~~{\eqref{eq:no_x}}
\State $\kappa^{(m)} = {\color{blue}\texttt{$\kappa$-noPBS}}(Y^{(m)}, \mathbf{B}^{(m-1)}, \zeta_{m}, \Theta)$~~~~~~~~~~~~~{\eqref{eq:no_kappa}}
\Else
\State $Z^{(m)} = {\color{blue}\texttt{Z-PBS}}(\mathbf{B}^{(m-1)}, \zeta_{m}, \Theta)$~~~~~~~~~~~~~~~~~~~~~~~~~~{\eqref{eq:pbs_z}}
\State $X^{(m)} = {\color{blue}\texttt{X-PBS}}(\mathbf{B}^{(m-1)}, \zeta_{m}, \Theta)$~~~~~~~~~~~~~~~~~~~~~~~~~~{\eqref{eq:pbs_x}}
\State $\kappa^{(m)} = {\color{blue}\texttt{$\kappa$-PBS}}(\mathbf{B}^{(m-1)}, \zeta_{m}, \Theta)$~~~~~~~~~~~~~~~~~~~~~~~~~~{\eqref{eq:pbs_kappa}}
\EndIf

\EndFor
\end{algorithmic}
\end{algorithm}
Algorithm~\ref{alg:mutual_info_planning} describes the implementation details of the proposed \textsc{$\kappa$-BNITP}. This algorithm is structurally similar to~\cite[Algorithm 1]{Srivastava.etal.RAL23}. In Algorithm~\ref{alg:mutual_info_planning}, the incorporation of the additional domain knowledge of the correlation between hazards and targets ($\kappa$-correlation) is highlighted in \textcolor{blue}{blue}. The algorithm begins by computing an informed path for the agent based on the current Bayesian network representation of the environment, using the subroutine \texttt{CalculateBNPath} (line 1), as described in~\cite[Algorithm 2]{Srivastava.etal.RAL23}. The output of the path-based sensor $\Theta$ is then observed, and if $\Theta=0$, the sensor readings are stored in $Y^{(m)}$. The posterior belief map is then computed using the corresponding function based on equations~\eqref{eq:no_z}, \eqref{eq:no_x}, and \eqref{eq:no_kappa}(line 6--8). In cases where the path-based sensor is triggered, i.e., $\Theta=1$, the sensor readings $Y^{(m)}$ are not available, and the posterior belief map is computed using the corresponding function based on equations~\eqref{eq:pbs_z}, \eqref{eq:pbs_x}, and \eqref{eq:pbs_kappa}(line 10--12). It must be noted that the subroutine \texttt{calculateBNPath} presented in Algorithm 1 extends the function \texttt{calculateBNPath}~\cite[Algorithm~2]{Srivastava.etal.RAL23} by incorporating the $\kappa$-correlation.

\section{Experiments And Results}
\label{sec:experiments_and_results}
\begin{figure*}[!t]
    \centering
    \includegraphics[width=\textwidth]{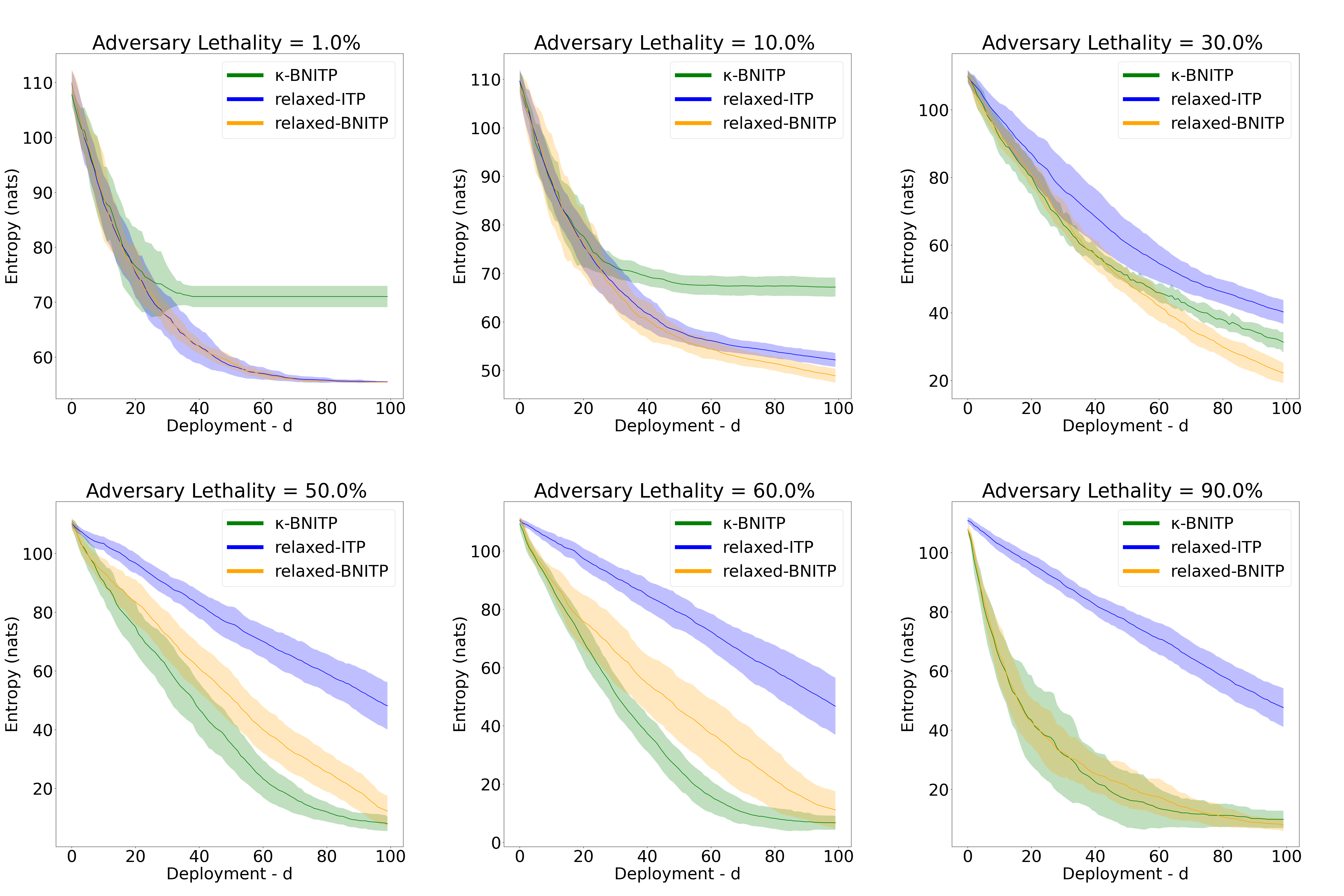}
    \caption{ Comparison of information entropy (in nats) 
    for \textsc{$\kappa$-BNITP}, \textsc{relaxed-ITP}, and \textsc{relaxed-BNITP} methods in environments with different adversary lethalities.
    }
    \label{fig:information_experiments}
\end{figure*}
This section presents the numerical experiments carried out to assess the efficacy of the proposed algorithm. Specifically, we compare \textsc{$\kappa$-BNITP} with~\cite{Otte.Sofge.TASE21} and \cite{Srivastava.etal.RAL23} in a spatial environment that includes hazards~$Z$ and targets~$X$. We begin by describing the experimental environment, followed by a discussion of the outcomes of the experiments.

\textbf{\emph{Experimental Setup}}: We conduct a series of $25$ Monte Carlo numerical experiments to evaluate the efficacy and robustness of the proposed \textsc{$\kappa$-BNITP} algorithm. The experiments are carried out in a discrete spatial environment with dimensions of $a\times b = 9\times9$ cells. Each cell in the environment could either be empty, contain hazardous elements ${\mathbf{Z}}$, which could lead to the destruction of the agent, contain targets $\mathbf{X}$, or contain both hazards and targets. The agents are deployed sequentially, and their movement in the environment is restricted to $9$-grid connectivity, allowing each agent to select the next step either by transitioning to any of its $8$-neighboring cells or by remaining in the same cell. To investigate how the hazard lethality, which refers to the likelihood of an agent's destruction if it reaches a cell $C$ that is occupied by a hazard ($Z_C = 1$), affects the information gathering process, we consider environments with hazard lethality of $1\%$, $10\%$, $30\%$, $50\%$, $60\%$, and $90\%$. Each agent has a malfunction probability of $5\%$ when taking a step, corresponding to the false-positive per step of a path-based sensor triggering. The likelihood of an agent surviving a cell $C$ with a hazard ($Z_C = 1$) is $5\%$, corresponding to the false-negative per step of a path-based sensor. The target sensor was assumed to truly detect targets in a cell $95\%$ of the time and has a $5\%$ false-positive rate of detection of targets.

\textbf{\emph{Results and Discussion}}: We conduct a comparative study of our proposed \textsc{$\kappa$-BNITP} algorithm with existing approaches, namely~\textsc{relaxed-ITP}~\cite[Algoirthm 1]{Otte.Sofge.TASE21} and~\textsc{relaxed-BNITP}\cite[Algoirthm 1]{Srivastava.etal.RAL23}. We deploy a fleet of $M=100$ agents sequentially and monitored the progress of Shannon information entropy (a measure of uncertainty) through each deployment, as shown in Fig.\ref{fig:information_experiments}. Our results demonstrate that \textsc{$\kappa$-BNITP} outperforms \textsc{relaxed-BNITP}, and \textsc{relaxed-ITP} only at a moderate lethality of $50\%$ and $60\%$. We observed that when the adversary lethality is higher, the algorithm underperforms and falsely reports the presence of targets in cells with only hazards. This bias can be attributed to the unreliable information obtained through the path-based sensors and the Bayesian network approach that relies on the accuracy of domain knowledge. As the path-based sensor observations are vague and incomplete, our attempt to infer the true state based on these observations is limited, leading to biases in lethality levels both lower and higher than $60\%$. Our results demonstrate that incorporating additional domain knowledge of $\kappa$-correlation helps improve the efficacy of information gathering. However, further research is needed to develop an accurate method 
that accurately estimates the belief map of the environment.
\section{Conclusion}
\label{sec:conclusion}

In this paper, we extend~\cite{Srivastava.etal.RAL23} and present a new algorithm called \textsc{$\kappa$-BNITP} for efficient and robust information gathering in a communication-denied environment with hazardous and target elements. The algorithm leverages the additional knowledge of the correlation between hazards and targets, referred to as $\kappa$-correlation, to improve the information gathering of target and hazardous elements. Our numerical experiments demonstrated the superiority of the proposed algorithm compared to existing approaches, namely \textsc{relaxed-ITP} and \textsc{relaxed-BNITP}, in terms of information entropy reduction at moderate hazard lethality levels. However, the algorithm underperforms at higher and lower hazard lethality levels due to the unreliable information obtained through path-based sensors and the Bayesian network approach that relies on the accuracy of domain knowledge. Overall, our results suggest that incorporating additional domain knowledge such as $\kappa$-correlation can improve the efficacy of information gathering, and future research is needed to develop more accurate models and approaches to estimate the environment.

\bibliographystyle{IEEEtran}
\bibliography{IEEEabrv,mybib}

\end{document}